\documentclass[jou,apacite]{apa6}

\usepackage{graphicx}
\usepackage[normalem]{ulem} 
\usepackage{amsmath}
\usepackage{xcolor}
\usepackage{url}
\usepackage{enumitem}
\usepackage{siunitx}    
\title{Hierarchical Shift Mixing – Beyond Dense Attention in Transformers}

\shorttitle{Forchheimer - Hierarchical Shift Mixing}

\author{Robert Forchheimer}
\affiliation{Dept. of Electrical Engineering, Linköping University, Linköping, Sweden\\RISE Research Institutes of Sweden, Linköping, Sweden}
\date{\today}

\keywords{LLMs, attention, token mixing, linear complexity, hierarchical, convolution}

\abstract{
Since the introduction of the Transformer architecture for large language models, the softmax-based attention layer has faced increasing scrutinity due to its quadratic-time computational complexity.  Attempts have been made to replace it with less complex methods, at the cost of reduced performance in most cases.
We introduce Hierarchical Shift Mixing (HSM), a general framework for token mixing
that distributes pairwise token interactions across Transformer layers rather than
computing them densely within each layer. HSM enables linear-time complexity while
remaining agnostic to the specific mixing function. We show that even simple HSM
variants achieve performance close to softmax attention, and that hybrid
architectures combining HSM with softmax attention can outperform a GPT-style Transformer baseline
while reducing computational cost during both training and inference.}

\begin{document}
\maketitle

\section{1. Introduction}
The \textit{Transformer} architecture was proposed in 2017 \cite{vaswani2017attention} as an improvement of earlier architectures for machine language translation. Consisting of an Encoder, driven by the first language, an intermediate representation of the input sentence is used by a Decoder to produce a corresponding translated sentence in the second language. With the ability to generate text, the decoder quickly became popular on its own as a large language model (LLM) to produce long sequences of text based on a seed (\textit{prompt}) given by the user. Central to both the encoder and the decoder is the  \textit{attention network} which ensures that the grammar as well as the semantics of the produced text are consistent. The attention network includes two basic mechanisms: a mixing part that combines words, and trained artificial neural networks, the purpose of the latter is to support the generation of the next word. For this reason the decoder architecture is also denoted as a \textit{Generative Pre-trained Transformer} (GPT). The next section presents the GPT architecture in more detail, and in particular its mixing part. The rest of the paper introduces Hierarchical Shift Mixing (HSM), a framework that offers linear-time complexity and is shown to perform well both when it comes to objective as well as subjective performance. HSM allows different token mixing functions and we show how they measure up against each other and to the GPT version. Finally, we show that a hybrid architecture consisting of a combination of GPT mixing and HSM mixing in different Transformer layers obtains better results than GPT mixing alone.

\section{2. The GPT Architecture}
\begin{figure}
  \includegraphics[width=\linewidth]{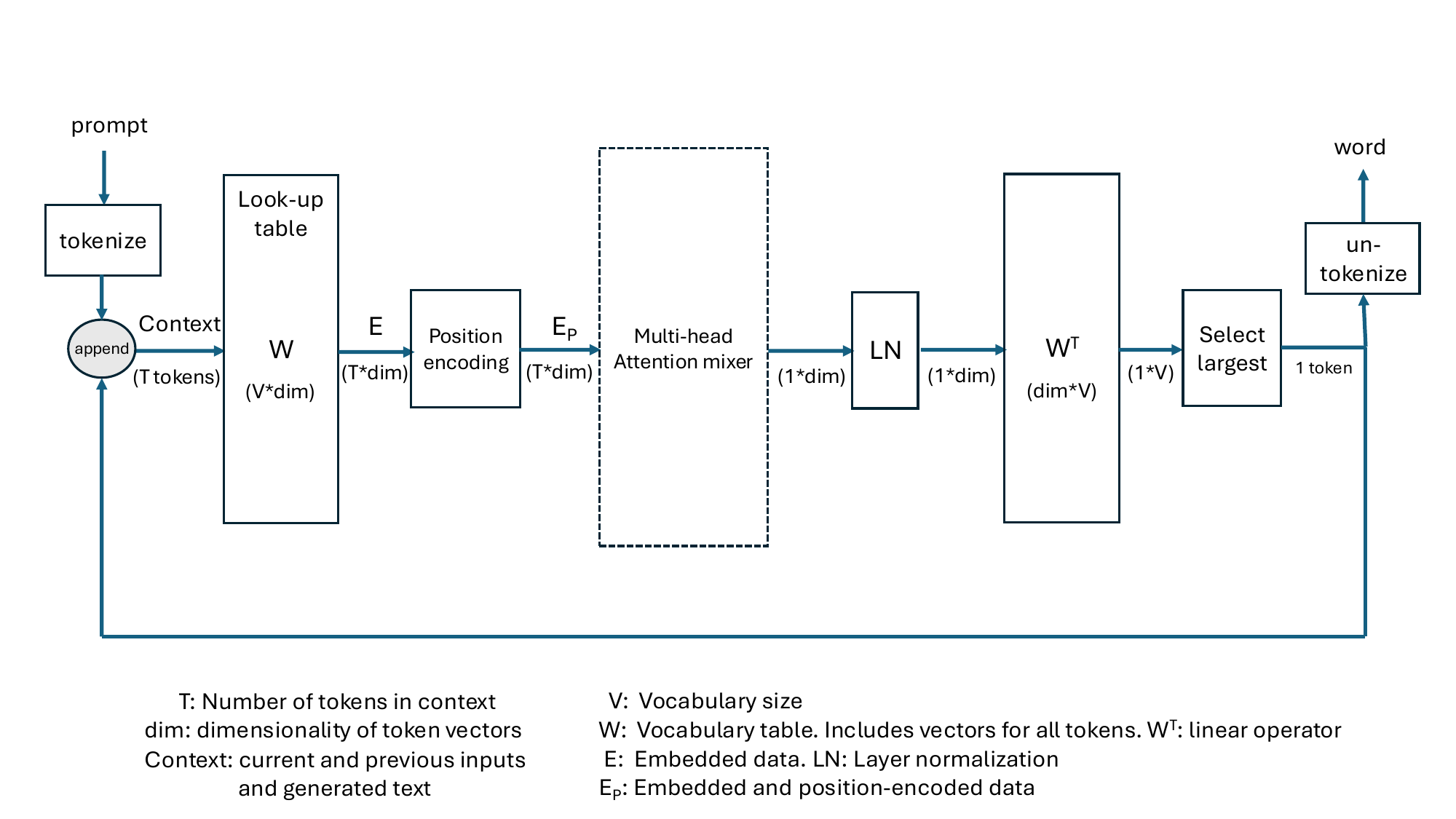}
     \vspace{-3pt} 
  \caption{Overview of the GPT architecture}
  \label{fig:gpt_architecture}
\end{figure}

Figure \ref{fig:gpt_architecture} gives an overview of the GPT architecture. The individual words of the user prompt are first split into tokens\footnote{\textit{Tokens} are typically parts of words. Also delimiters become tokens.},  which are then converted to vectors (\textit{embedding}). Although tokens do not necessarily correspond to whole words, we use the terms interchangeably for simplicity.
 The vector components are either precomputed or learned by training them together with other parameters of the model. We denote the dimensionality of these vectors $dim$.
The prompt is held in the \textit{context window} which is large enough to also store subsequently generated words and additional user inputs. This allows the algorithm to have access to the recent dialogue. The group of vectors corresponding to the content of the context window is then augmented, based on the position of the corresponding words in the window. The augmented vectors are further processed in the \textit{Multi-head Attention Mixer} to generate a \textit{contextual vector} that represents one, or several, possible predicted words. Following normalization (LN), the same embedding table as on the input side is used, but now in a reversed fashion by (effectively) searching for the nearest word that corresponds to the contextual vector\footnote{Using the input embedding table instead of a separately generated table is denoted \textit{tied embedding} and has turned out to work well in the Transformer scheme.}. In practice, there may be several candidate words whose vectors are close to the contextual vector. Similarity can be measured through the dot product, which can be interpreted as a (non-normalized) correlation between vectors. As indicated in Figure \ref{fig:gpt_architecture}, the most similar vector (giving the largest dot product) is a good candidate. However, depending on the setting of a user-controllable parameter (\textit{temperature}) one of the other likely candidates can be randomly chosen as the next generated word. In this way, the algorithm becomes less deterministic and allows different  outcomes to be generated even when initiated with the same prompt.

\subsection{2.1 The GPT mixer}
\begin{figure}[htbp]
 \centering
  \includegraphics[width=\linewidth]{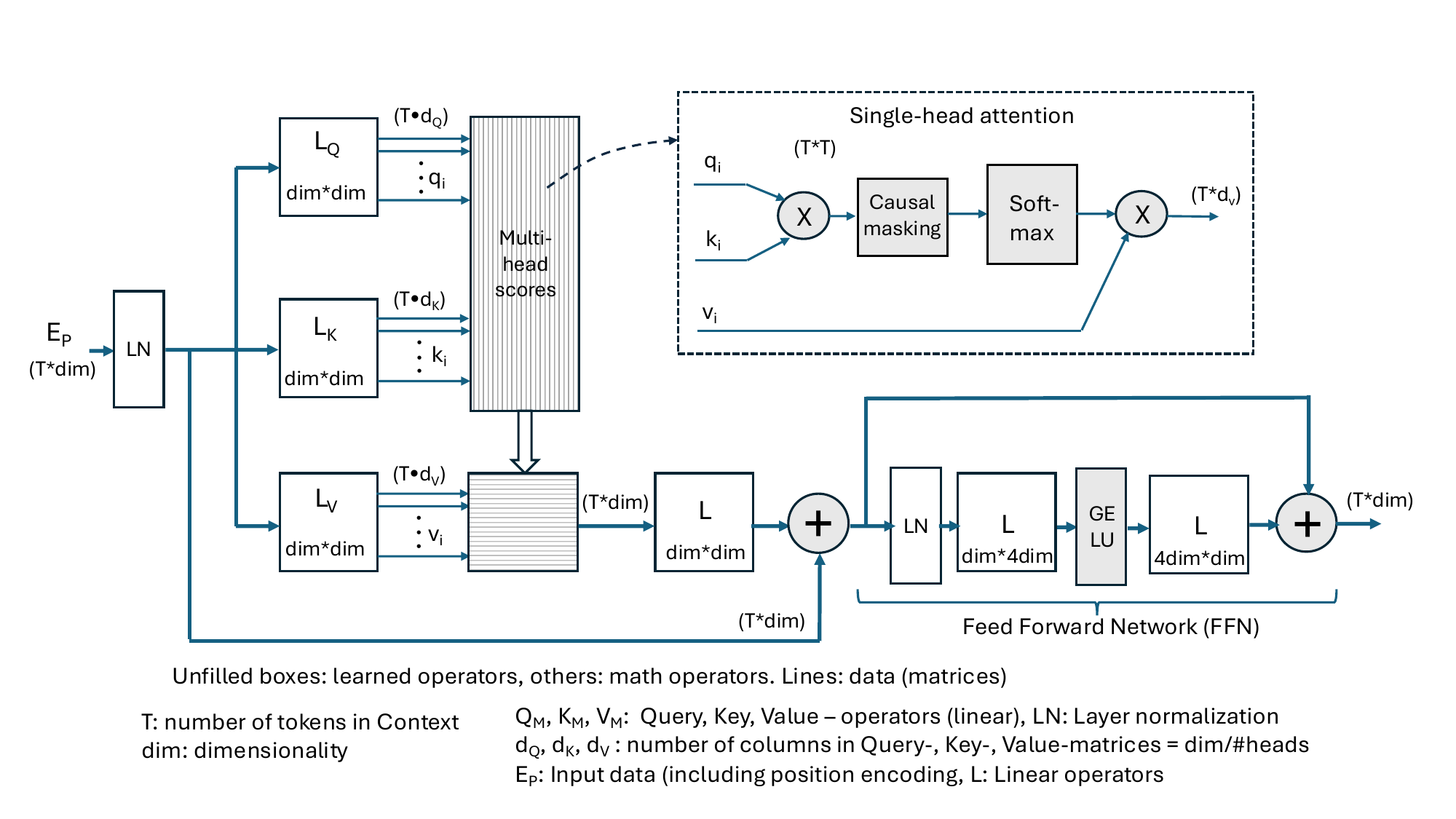} 
     \vspace{-3pt} 
  \caption{GPT mixer – one layer.}
  \label{fig:gpt_mixer}
\end{figure}

\begin{figure}[htbp]
 \centering
  \includegraphics[width=\linewidth]{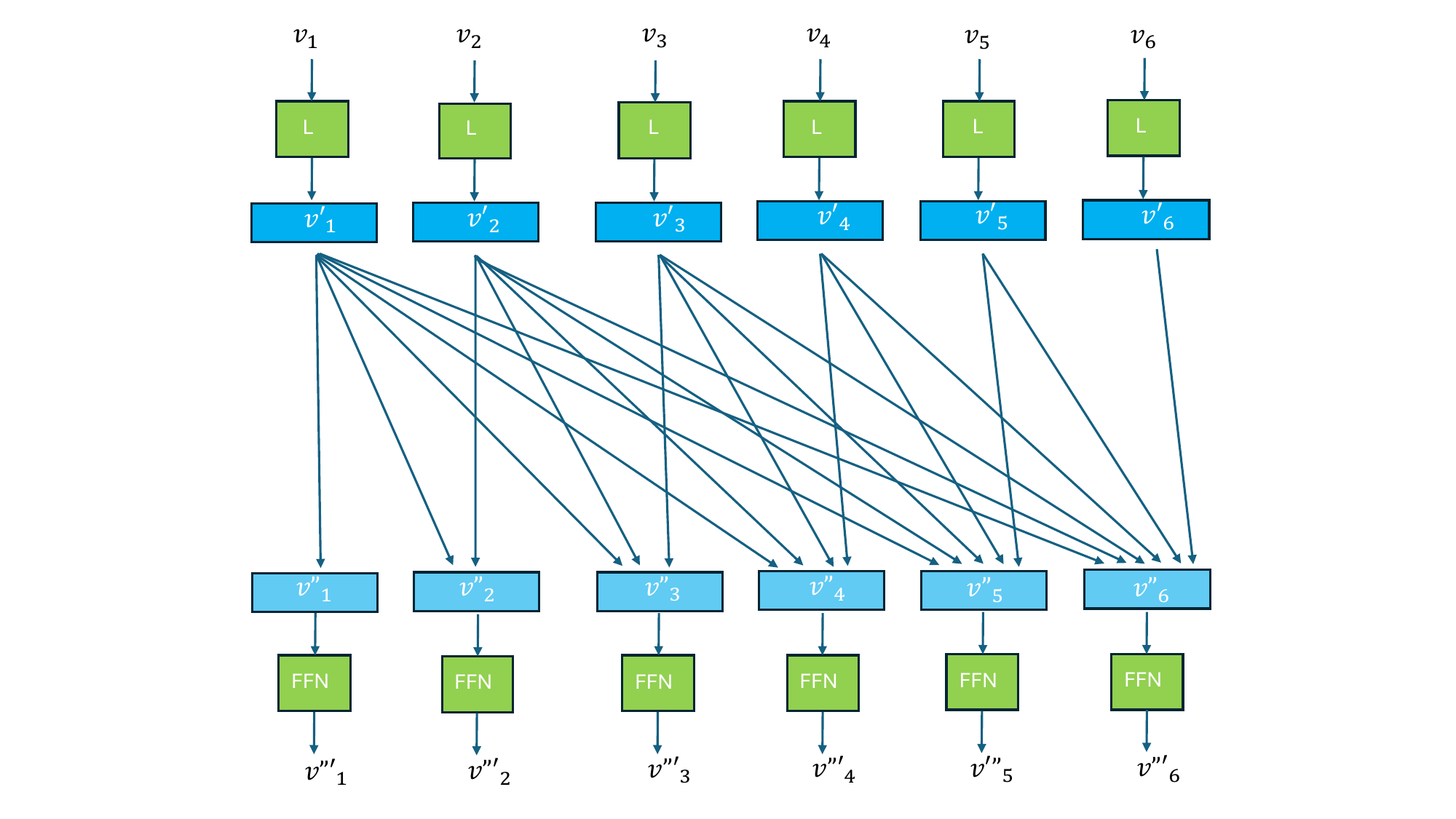} 
     \vspace{-3pt} 
  \caption{How tokens are mixed in each GPT attention layer.}
  \label{fig:gpt_attention}
\end{figure}

At the heart of the GPT mixer lies the attention mechanism. It plays several roles. Based on the word vectors that correspond to the words in the context window, it modifies each vector to better align with the semantic meaning of the other words. As an example, the word \textit{bank} may refer to a financial institution or to a river bank\footnote{Such words are referred to as \textit{homonyms.}}. The vector representation may initially include both meanings, in the sense that  vector components reflecting, e.g., "wetness" and "money" both have high values. If the context contains the sentence "\textit{Because of flooding, the bank was closed}" the attention mechanism will modify the vector representation of bank to move towards a representation of \textit{river bank}.
Another task for the attention mechanism is to connect words which are related, even if they are far from each other in the text. For example, in the sentence "\textit{Peter put on his coat and shoes and went to the car. He then drove out on the street...}", the words \textit{Peter} and \textit{He} refer to the same entity, which needs to be understood by the system.
The GPT mixer consists of repeated attention layers, where the detailed functionality of each layer is shown in Figure \ref{fig:gpt_mixer}. An illustration of the token mixing within one layer is shown in Figure \ref{fig:gpt_attention} for a sentence containing 6 words.

Each word is augmented by linearly combining it with all \uline{previous} words in the context window. The  weights for a target word are found through dot products between the target word and each of the previous words. These dot products are then normalized using the softmax function resulting in the final weight values. For this reason, the GPT attention method is sometimes also referred to as \textit{softmax attention}. We will alternatively use the term \textit{dense attention} emphasizing that all previous tokens contribute to the
update. 
The attention layer is preceded by a linear mapping (\textit{L}) of each word and followed by small neural networks (Feed Forward Networks, \textit{FFNs}) at the output. Finally, a number of these layers are stacked on top of each other to gradually modify all word vectors. The rightmost word vector from the last layer is selected as the output from the mixer. Thanks to the training of the \textit{L} and \textit{FFN} parameters, this vector will usually produce a good prediction of the next word. It is worth noting that all \textit{L} mappings share the same  values within the same layer. This is also true for the \textit{FFN}s. It should be noted that Figure \ref{fig:gpt_attention} has been simplified as \textit{L} in this figure corresponds to the $L_V$ (Value) mapping in Figure \ref{fig:gpt_mixer} while $L_Q,$ (Query) and  $L_K$ (Key) projections, illustrated in Figure \ref{fig:gpt_mixer}, that precede the dot products are not shown. In addition, the residual connection into the \textit{FFN} input as well as multihead attention are omitted in Figure \ref{fig:gpt_attention}. We will return to these issues later on.

\section{3. Hierarchical Shift Mixing}
\begin{figure}[htbp]
   \includegraphics[width=\linewidth]{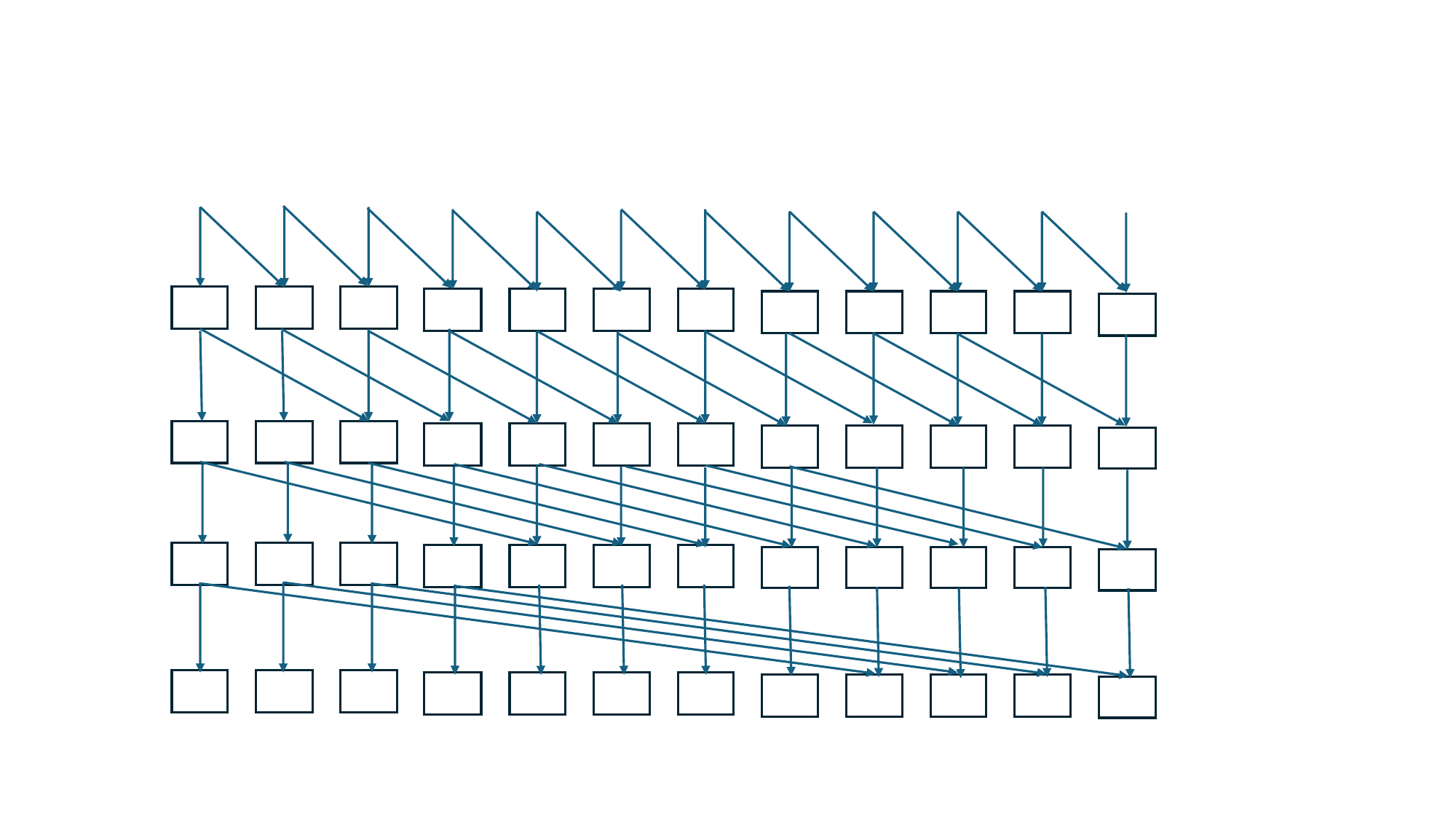} 
  \caption{Hierarchical shift mixing showing a stack of 4 layers.}
  \label{fig:hsm_mixing}
\end{figure}

GPT mixing computes pairwise dot products between all words and their preceding words. With $T$ words in the context window this amounts to $O(T^2)$ interactions. Thus, the computational complexity grows quadratically with the number of words in the context window. There are several proposals in the literature of other mixers that reduce the complexity to $O(Tlog(T))$ or even $O(T)$, the latter corresponding to linear-time complexity. A common denominator is that these methods usually bring with them a reduction in performance. \\
\indent In this paper we introduce a general architecture, Hierarchical Shift Mixing (HSM), where the focus is not on the actual mixing performed on the token pairs, but on the way these pairs are selected in order to implement alternatives to attention-based mixing.\footnote{\textit{Attention} in the context of LLMs usually denotes the specific function described in \cite{vaswani2017attention}. We will respect this practice and use  \textit{token mixing} for our broader class of functions.}
Instead of processing all pairs at each layer, a subset of the pairs is selected in such a way that all pairs will be addressed across the full layer stack.\\ 
\indent Figure \ref{fig:hsm_mixing} illustrates this principle. In the first (top) layer, only nearby pairs are processed, in the next layer, pairs at a distance of two are processed, next pairs at a distance of four etc. As can be seen, causality is preserved by this construction as no future tokens are involved. Thus, HSM fully preserves the parallelism and efficiency of the transformer scheme.
Each box in Figure \ref{fig:hsm_mixing} represents a possible token mixing function to be performed on a selected pair. This function can range from a simple weighted summation, to nonlinear dynamic gating, or to a full neural network. The variations that we have studied are described in the following sections and are further illustrated in Figures \ref{fig:linattention} and \ref{fig:nonlinattention}. The two (vector) inputs to each box are denoted $\mathbf{x}$ and $\mathbf{x}_{\text{shifted}}$ respectively and the output is denoted $\mathbf{y}$. In the case where there is only one input, then $\mathbf{x}_{\text{shifted}} = 0$. Similar to the GPT scheme, the $\mathbf{y}$ output is further processed by a neural network (FFN) before entering the next layer (implicit in Figure \ref{fig:hsm_mixing}). 

\subsection{3.1 ({$a, b$}) weighting}

Our first token mixing function simply adds the two incoming vectors, weighted by scalar weights $a$ and $b$ respectively:

\begin{equation}
 \mathbf{y} = a\mathbf{x} + b\mathbf{x}_{\text{shifted}}
 \end{equation}

The values of $a$ and $b$ are learned in the training phase and stay the same within each layer but differ between  layers. The operation is illustrated in Figure \ref{fig:linattention}a.

\subsection{3.2 ($\mathbf{a}, \mathbf{b}$) vector weighting}
More flexibility is obtained by applying different weights to each feature of the word vectors. This is done through \uline{element-wise} multiplication by learned coefficient vectors $\mathbf{a}$ and $\mathbf{b}$ with the two input vectors respectively:

\begin{equation}
\mathbf{y} = \mathbf{a} \odot\mathbf{x} + \mathbf{b}\odot\mathbf{x}_{\text{shifted}}
\end{equation}

As in the previous case, $\mathbf{a}$ and $\mathbf{b}$ stay the same within each layer but differ between layers. Figure \ref{fig:linattention}b illustrates this case.

\subsection{3.3 (A, B) weighting}
\begin{figure}[htbp]
   \includegraphics[width=\linewidth]{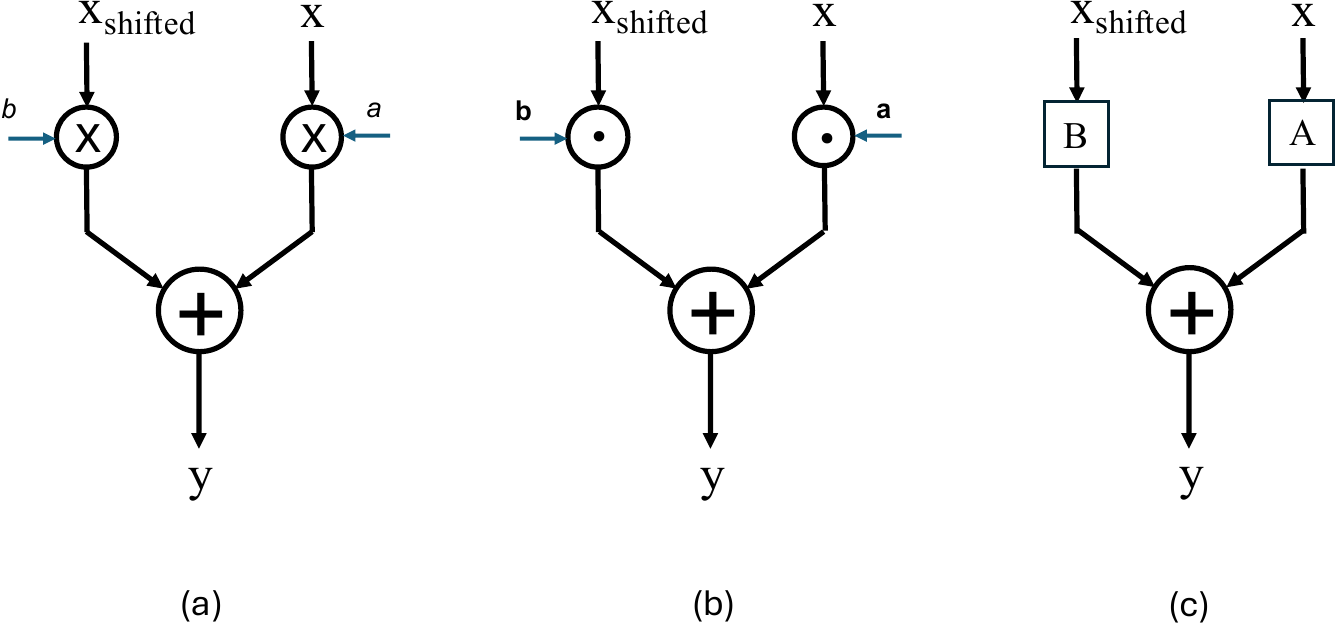} 
   \vspace{3pt} 
  \caption{Linear mixing functions.}
  \label{fig:linattention}
\end{figure}

As shown in Figures \ref{fig:gpt_mixer} and \ref{fig:gpt_attention}, the GPT attention scheme applies a linear projection on each word vector $v_i$ before weighting them together to form $v''_i$. The weighting is computed from two other linear mappings, namely  $L_Q$ when the output of $v_i$ corresponds to $\mathbf{x}$ and and  $L_K$ when the output of $v_i$ corresponds to $\mathbf{x}_{\text{shifted}}$.  In our (A, B) weighting scheme (see Figure \ref{fig:linattention}c), we inherit a similar idea, replacing $L_V$ with two learned mappings  A and B respectively. (A,B) weighting, together with a bias term, represents the most general way to combine the two vectors linearly:

\begin{equation}
\mathbf{y} = A\mathbf{x} + B\mathbf{x}_{\text{shifted}} + \textit{bias}
\end{equation}

\subsection{3.4 Nonlinear weighting} 
Our remaining HSM schemes combine the two input vectors in nonlinear ways. This allows the words in the context window to dynamically influence the weights. Similar to the earlier linear methods, these approaches still have linear-time complexity.
Here we consider three different cases with increasing expressive power, which we denote as: i) gated weighting with single input, ii) gated weighting with double input, and iii) fusion. \\
All of these methods are characterized by dynamic (inference-time) modifications of the weights. Although less common in the context of neural networks, similar mechanisms  are found in many architectures, such as  GPT attention as well as in networks for computer vision \cite{sabour2017dynamic}, \cite{pmlr-v97-ratzlaff19a}. 

\subsection{3.5 Gated weighting - single input} 
In the first version of gated weighting, the input $\mathbf{x}$ is used to influence the weighting as follows (see also Figure \ref{fig:nonlinattention}a): 

\begin{equation}
\begin{aligned}
&gate = \tanh(\mathrm{mlp}(\mathbf{x})) \\
&\mathbf{y} = gate \odot \mathbf{x} + (1 - gate) \odot \mathbf{x}_{\text{shifted}}
\end{aligned}
\end{equation}

\noindent where  \textit{tanh} is a nonlinear function\footnote{$
\tanh(x) \coloneqq \frac{e^x - e^{-x}}{e^x + e^{-x}}$}, and \textit{mlp} denotes a multi-layer perceptron implemented as a minimal two-layer neural network (Linear($dim$→$dim$) → ReLU → Linear($dim$→$dim$)). The use of \textit{tanh} constrains the weights to the range [-1,1], which helps stabilize the learning process. Gating with \textit{tanh} is also found in other neural network architectures, such as recurrent neural networks and LSTMs \cite{hochreiter1997long}.

\subsection{3.6 Gated weighting - double input}
Using both $\mathbf{x}$ and $\mathbf{x}_{\text{shifted}}$ extends the gated weighting further. The vectors are concatenated and passed through a simple neural network that produces the gating via a linear mapping $(L)$ followed by the \textit{tanh} function (see  Figure \ref{fig:nonlinattention}b): 

\begin{equation}
\begin{aligned}
&gate = \tanh\!\big(L(\mathrm{concat}(\mathbf{x}, \mathbf{x}_{\text{shifted}}))\big) \\
&\mathbf{y} = gate \odot \mathbf{x} + (1 - gate) \odot \mathbf{x}_{\text{shifted}}
\end{aligned}
\end{equation}

\noindent Similar to single-input gated weighting, the double-input version is content-aware. In this case, the blend depends on both current and past features, and cross-channel interactions are enabled by the linear projection and the FFN layer.

\subsection{3.7 Fusion}
\begin{figure}[htbp]
   \includegraphics[width=\linewidth]{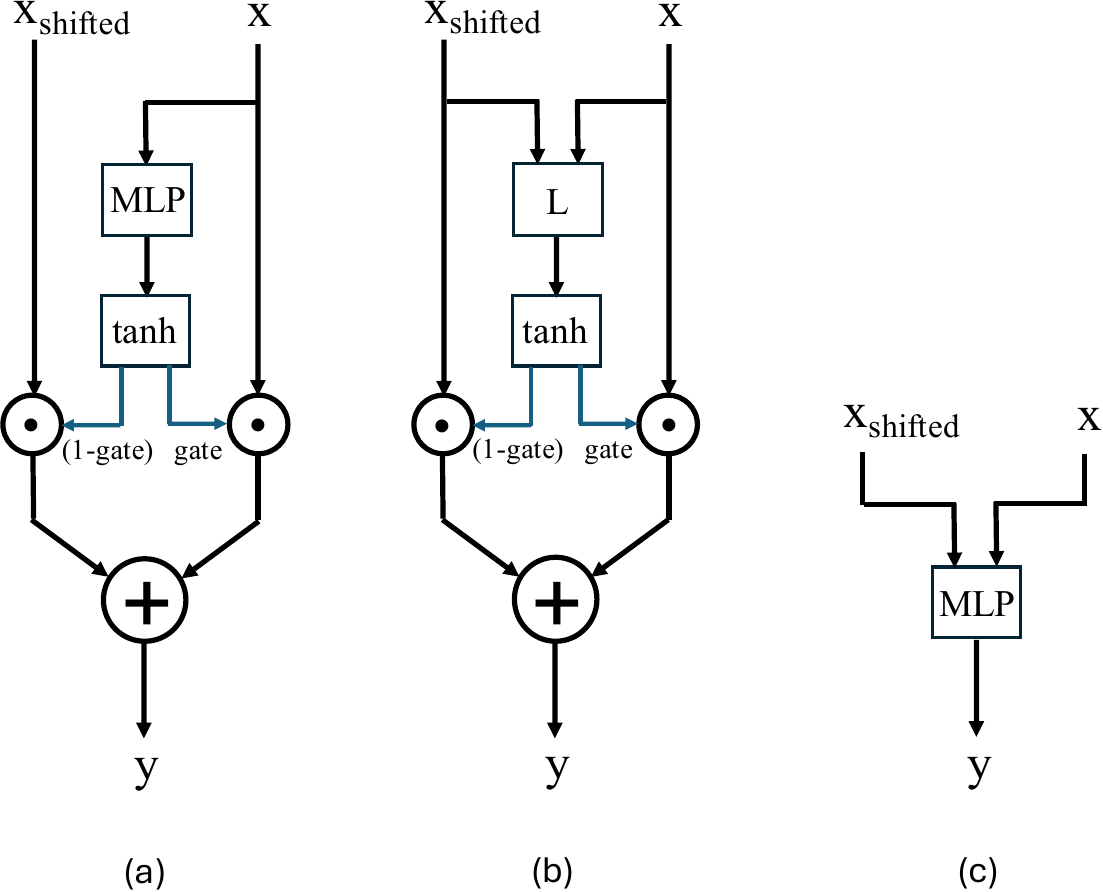} 
   \vspace{3pt} 
  \caption{Non-linear mixing functions.}
  \label{fig:nonlinattention}
\end{figure}

The fusion model starts in an identical way to the double-input gating method. However,  the nonlinear gating function is removed, and the neural network directly produces the new features (see Figure \ref{fig:nonlinattention}c):

\begin{equation}
\mathbf{y} = \mathrm{mlp}(\mathrm{concat}(\mathbf{x}, \mathbf{x}_{\text{shifted}}))
\end{equation}

This formulation is no longer restricted to linear or signed weighting of $\mathbf{x}$ and $\mathbf{x}_{\text{shifted}}$. Instead, the neural network can synthesize features from both inputs. In this case we use a three-layer network (Linear(2$\cdot head\_dim$→$head\_dim$) → ReLU → Linear($head\_dim$→$head\_dim$)) where $head\_dim$ is the dimensionality of one head (see next section). The fusion mechanism brings us close to the full functionality of GPT attention still without imposing quadratic-time complexity.

\subsection{4. Multihead HSM}
A fair comparison with GPT attention would not be possible without taking \textit{multihead attention} into account. In the GPT version, this amounts to dividing up the features of the word vectors into $n$ groups containing $dim/n$ features each. Each group runs its own attention process using correspondingly reduced-size key and query mappings. Before reaching the FFNs, the results from all the $n$ groups are concatenated back into a full-sized ($dim$) vector. Multihead attention allows each word to test several alternative ways of relating to other words, thus reducing the risk of committing to a single interpretation. It increases expressive power and has been shown to improve performance in Transformer architectures.
Some of the proposed HSM schemes lend themselves to multihead implementation, while others, such as ($\mathbf{a}, \mathbf{b}$) weighting, already include the flexibility to process different features independently. As an initial attempt we apply the multihead concept to the gated weighting - double input scheme as well as to the fusion scheme. In these cases, each head within a layer uses the same ($\mathbf{x}$, $\mathbf{x}_{\text{shifted}}$) pair. The distance between the pairs doubles at each layer according to the HSM principal. With several heads, the number of linear and non-linear mappings increases by the number of heads per layer. However, due to the reduced number of features per head, the total number of parameters decreases. Each head is less expressive individually while overall capacity may increase due to diversity.
Furthermore, with multiple heads there are more opportunities. Not only can the mixing differ between groups of features, but each head can also use its own shift pattern. We apply this idea to the ($a, b$) scheme. In  "HSM ($a, b$) Multihead" version, the first head vill use a shift of 1 step, the second a shift of 2, the third a shift of 4 and so on. This means that each layer processes multiple time shifts, in contrast to the earlier versions where each layer processed only a single shift. This brings the model closer to GPT attention while retaining linear-time complexity, since not every token is mixed with every other token in each layer. In this version, all layers use the same shift pattern.

A final multihead extension, "HSM ($a, b$) Multihead-ext", is introduced in section 7. Here, different time shifts are used in different layers. This represents the closest HSM approximation to full attention while preserving linear-time efficiency.

\section{5. Hybrid HSM-GPT mixing}
In GPT, the attention layers operate independently of each other, and their input data format is identical to their output format, namely the number of words in the context window multiplied by the size of the embedding vectors. We adopt the same concept for HSM. This opens up the possibility of applying different mixing schemes in different layers. Such hybrid techniques have been reported earlier, see e.g. \cite{YunBRRK20}, \cite{liu2021pay}, \cite{tay2021synthesizer},  \cite{choromanski2021rethinking}, \cite{wang2020linformer} and have been shown to provide advantages in performance and/or computational complexity. Since HSM has linear-time complexity, replacing some of the GPT layers in the GPT architecture with HSM layers will reduce the overall computation cost. The key question is how this replacement affects performance. In section 7, we show that certain combinations of HSM and GPT layers not only achieve comparable training loss to the GPT reference model, but can also surpass its performance.\\

\section{6. Evaluation}
Evaluating the various schemes requires that we specify the training data, the details of the algorithms, including  hyperparameters, and how performance is measured. 

\subsection{6.1 Transformer implementations}
Shortly after the publication of the Transformer paper by Google in 2017, improvements were reported by different academic and industrial groups. In particular, the release in 2019 of the GPT-2 language model\footnote{The prefix "Chat" was added after the model was trained on a large number of question-answering examples (\textit{fine tuning}) to behave as a conversational tool.} \cite{radford2019language} included several modifications. One of these showed that normalization of the intermediate results in each attention layer could advantageously be applied to the input rather than the output (cf. Figure \ref{fig:gpt_mixer}). This approach is denoted \textit{pre-layer normalization} and replaces the \textit{post-layer normalization} scheme used in the original Transformer. GPT-2 also includes a final layer normalization (LN) block before the linear output mapping (see Figure \ref{fig:gpt_architecture}). Finally, GPT-2 uses learned positional encodings instead of fixed (sinusoidal) positional encodings.

\paragraph{The GPT reference}
Our  reference model is denoted simply as $GPT$ and is modeled after GPT-2. To enable training in a reasonable time on a desktop computer, the model has been scaled down substantially. The dimensionality of the embedding vectors (\textit{dim}) has been set to 256 (768 for the original GPT-2 Small configuration) and the size of the context window is 128 (1024) tokens. The vocabulary size is 5000 (50257), the FFN dimensionality is 512 (3072), the number of layers is 7 (12), and the number of attention heads is 8 (12). In total, the number of trainable parameters amounts to 5.1 million (124 million). 
\paragraph{HSM}
With reference to the general structure shown in Figure \ref{fig:gpt_architecture}, HSM replaces the GPT multihead attention mixer, with the main differences introduced by the various HSM token mixing schemes. Some of these schemes use linear projections with a relatively large number of parameters on the input side of each layer such as (A, B) weighting,
double-input gating, and fusion. In contrast, ($a,b$) weighting and ($\mathbf{a}, \mathbf{b}$) weighting use very few parameters at the input.
To keep the total number of parameters the same across schemes, the size of the FFN networks is adjusted accordingly.  The aim is to keep all proposed versions as close as possible to 5.1 million parameters enabling a fair comparison.

\subsection{6.2 Training data}
OpenAI used approximately 40 GB of text to train their GPT-2 model \cite{openai2019gpt2blog}. In this work, the substantially smaller dataset "TinyStories" is used \cite{eldan2023tinystories}. This dataset, which is available on the HuggingFace repository  \cite{tinystories-dataset}, has a size of 1.9 GB and consists of short stories written in a language corresponding to that of children aged 3-4 years. The data was divided into training data (90\%) and test data (10\%)\footnote{The actual training data size is even lower as stories shorter than the context window size (128 tokens) were filtered out.} and tokenized using a custom-trained byte-level BPE tokenizer \cite{wolf-etal-2020-transformers}.

\subsection{6.3 Objective measures}
During training, the unknown parameters are modified iteratively as new tokens from the training set enter into the context window. The goal of this process is to reduce, at each step, the error between the predicted next token and the true token. Starting at the output, a loss function ($\mathcal{L}$), describing the error as a continuous value, is computed. The parameters of the network are then
adjusted such that $\mathcal{L}$ is reduced. This procedure, known as backpropagation, is applied to all parameters in the model.

The loss function used in this work is the cross-entropy loss. It uses the output of the final state of the model (the normalized output of the linear mapping $W^{\mathrm{T}}$, in Figure \ref{fig:gpt_architecture}) to estimate a probability density \textit{q(x)} over all possible token predictions \textit{x}. This density function is compared to a (hypothetical) correct distribution \textit{p(x)} using the cross entropy:

\begin{equation}
H(p,q) = - \sum_{x \in \mathcal{X}} p(x) \log q(x).
\end{equation}

The cross entropy can be interpreted as the average uncertainty about the outcome of the next token when the distribution \textit{q(x)} is assumed while the correct distribution is \textit{p(x)}.

Since \textit{p(x)} is unknown, it is customary to replace it with a distribution concentrated on the correct token $x_{true}$. Then, the above expression reduces to:

\[
\mathcal{L} = - \log q(x_{\text{true}})
\]

\noindent which defines the cross-entropy loss during training. \\

In addition, we also estimate the probability of predicting the correct token ("validation accuracy"). We expect to see a strong (anti-) correlation between these two measures. When accuracy is high, loss should be low.

\subsection{6.4 Subjective (qualitative) evaluation}
We make a qualitative evaluation of the various mixing functions by analyzing their generated sequences when the models are given specially designed prompts. This is the same method that is used in  \cite{eldan2023tinystories} and we will follow their color coding, where a red background indicates poor semantic coherence (based on our qualitative judgment), yellow background indicates partial semantic coherence and green background indicates good coherence. An example of such a prompt is the following, \\

\textit{Alice was so tired when she got home so she went...}\\

where the generated completion of the sentence could be\\
\begin{itemize}[label={}, leftmargin=20pt, itemsep=0pt, parsep=0pt, topsep=0pt]
  \item...to her room. (red background)
  \item...to her bedroom. (yellow background)
  \item...to bed. (green background)
\end{itemize}

\section{7. Experimental results}
Our experiments aim to compare the performance of all HSM variants described earlier. In addition, we include GPT and two hybrid combinations of HSM and GPT. \\

The following hyperparameters are shared across all models: 
\begin{itemize}[label={}, leftmargin=20pt, itemsep=0pt, parsep=0pt, topsep=0pt]
\item- Batch size: 256
\item- Learning rate: 0.002
\item- Dropout rate: 0.1
\item- Epochs: 20\\
\end{itemize} 

The models are trained using the AdamW optimizer \cite{loshchilov2019adamw}. Validation loss and validation accuracy are recorded after each epoch to allow monitoring of model performance.

\paragraph{Qualitative results}
All models are able to produce grammatically correct text given an initial prompt. An example is shown below, where the short story was generated by the ($a, b$) model and the prompt is shown in bold:\\

{
\setlength{\parskip}{-3pt}   
\setlength{\parindent}{0pt} 
{\footnotesize\
\textbf{Once upon a time}, there was a little girl named Lily. She loved to play outside in the sunshine. One day, she saw a big, scary dog. The dog was barking and running towards her. Lily was scared and didn't know what to do.\\

Suddenly, a kind man came and asked Lily what was wrong. Lily told him about the dog and how she wanted to help him. The man smiled and said, "Don't worry, I'll help you." He took out a stick and gently cut the dog's fur. The dog was so happy and wagged his tail.\\

Lily and the man became friends and played together every day. They were the best of friends and always helped each other when they needed it. And they all lived happily ever after. The end.}
} \\

Although the semantic coherence is somewhat limited, the generated text is fairly consistent considering the limited context length of 128 tokens. Even the paragraph breaks were produced by the model.

Table \ref{tab:subjective} summarizes the qualitative semantic evaluation of the different models using the color-coding scheme described in the previous section. \\

\paragraph{Objective results}
Table \ref{tab:objective} summarizes the different HSM variants described above including the GPT reference model.
The performance (validation loss) of the pure HSM models is found in the upper part of this table.
The mixing functions range from the simplest ($a, b$) scheme, which performs a weighted sum using two scalar values, to methods that perform linear (A,B) and nonlinear (gating and fusion) operations as part of their mixing functions. Interestingly, the ($a, b$) function achieves the lowest validation loss among the pure HSM models (indicated in bold in Table \ref{tab:objective}). Table \ref{tab:ab_layers} shows an example of  the learned values of the weights $a$ and $b$ at each layer. Contrary to intuition, the updated value of $\mathbf{x}$ is almost entirely dominated by $\mathbf{x}_{\text{shifted}}$. However, it should be noted that this effect is partly offset by the residual signal path in which $\mathbf{x}$ is added to the mixed output.

\begin{table}[t]
\centering
\includegraphics[width=\columnwidth]{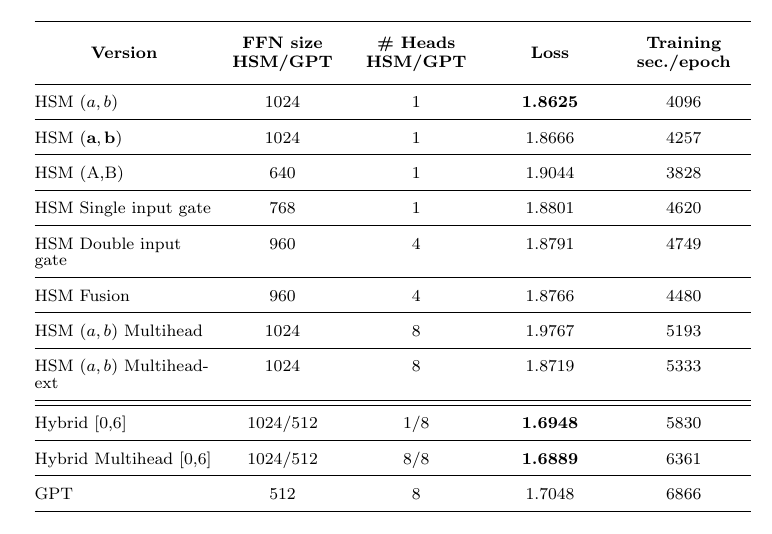} 
\caption{HSM configurations and results.}
\label{tab:objective}
\end{table}

More generally, the results suggest that allocating parameters to the FFN component is more beneficial than allocating them to linear or nonlinear pre-computations in the mixing function.

The multihead version of HSM ($a, b$) performs worse than its single-head counterpart. This is most likely due to the fact that all layers use the same shift structure, such that each head always sees the same shift distance and not all tokens pairs are covered across layers.

This limitation can be addressed by permuting the shift distances across layers. Using a rotating permutation scheme ([1,2,4,8,16,32,64,128] → [2,4,8,16,32,64,128,1] ...→ [64,128,1,2,4,8,16,32]) for the seven layers brings the scheme closer to the original HSM philosophy. Introducing this modification reduces the validation loss to 1.8719 (denoted "HSM ($a, b$) Multihead-ext" in Table  \ref{tab:objective}). While this is a substantial improvement over the original multihead variant, it still does not match the performance of the single-head version.
 
Table \ref{tab:objective} also includes results for two hybrid combinations as well as the GPT reference. "Hybrid [0,6]" replaces the first and last of the seven-layer GPT model with the corresponding layers from "HSM ($a, b$)" in these positions. Similarly, "Hybrid Multihead [0,6]" uses "HSM ($a, b$) Multihead" in these positions. Notably, although the multihead HSM model performs  worse than its single-head counterpart in isolation, it achieves the best performance when used in the hybrid configuration.

Both hybrid schemes outperform GPT, as indicated by the bold values in Table \ref{tab:objective}. It should be noted, however, that the hybrid models no longer have linear-time complexity.

\begin{figure}[htbp]
   \includegraphics[width=\linewidth]{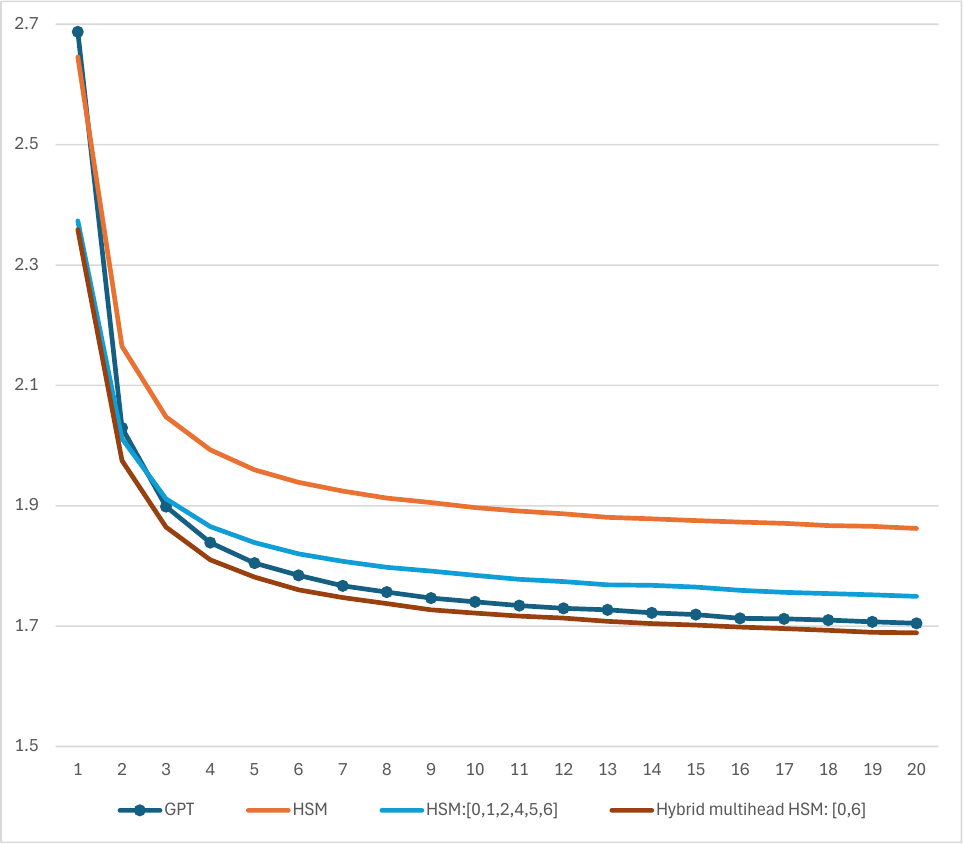} 
   \vspace{3pt} 
  \caption{Validation loss vs epochs. "HSM" designates the HSM ($a, b$) model,  "HSM:[0,1,2,4,5,6]" is a hybrid architecture where layer 3 has been replaced by softmax attention. Others as defined in Table \ref{tab:objective}}
  \label{fig:graphs}
\end{figure}

Figure \ref{fig:graphs} shows the training convergence of four of the models. "Hybrid [0,6]" is not  included as its curve coincides almost exactly with that of GPT. Of particular interest is that "Hybrid Multihead [0,6]" outperforms GPT while also converging faster during training.

\begin{figure}[htbp]
   \includegraphics[width=\linewidth]{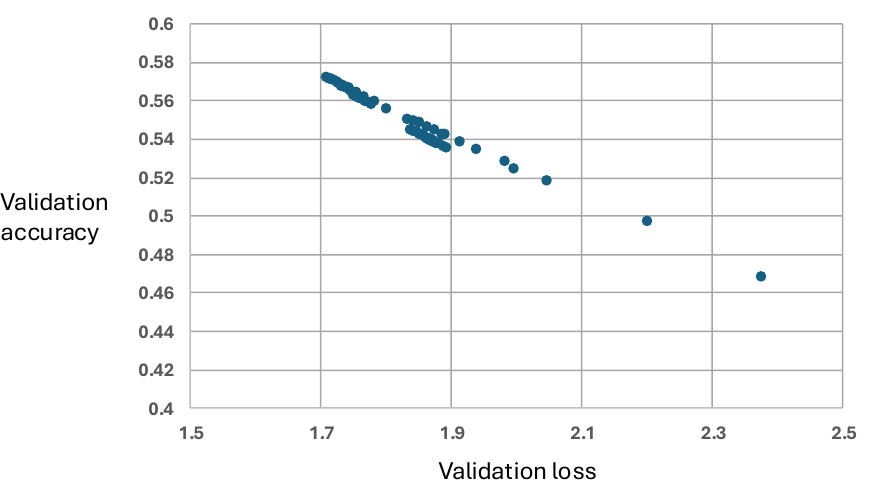} 
   \vspace{2pt} 
  \caption{Validation accuracy vs loss }
  \label{fig:accuracy}
\end{figure}

\paragraph{Validation accuracy vs loss}
Our second objective measure is validation accuracy. This is a measure of the fraction of correct next-token predictions during validation. In principle, this accuracy should be related to loss, and it is obvious from Figure \ref{fig:accuracy} that such a relation exists. Figure \ref{fig:accuracy} shows a point cloud of a large number of accuracy-loss pairs collected during validation for all models. There is very good agreement between increasing loss and decreasing accuracy.  A few datapoints deviate slightly from the regression trend. A closer inspection reveals that these points all originate from one model, namely the "HSM ($a, b$)" model. The reason for this behavior remains an interesting open question but has not been further investigated.

\paragraph{Training time vs performance}
Training times, measured in seconds per epoch, are reported in Table \ref{tab:objective}. As expected, all HSM variants train faster than the GPT model. The simplest version of HSM ($a, b$) is 40\% faster,  albeit with lower performance. The hybrid mode "Hybrid [0,6]", in which the first and last attention layers of the GPT are replaced by HSM ($a, b$) layers, achieves performance comparable to GPT while being 15\% faster. Finally, Hybrid Multihead [0,6] not only performs better than GPT, but is also 7.3\% faster.

\begin{table}[h!]
\centering
\resizebox{\columnwidth}{!}{%
\sisetup{
    table-number-alignment = center,
    table-figures-integer = 1,
    table-figures-decimal = 4
}
\begin{tabular}{l*{7}{S}}
\toprule
 & \textbf{Layer 0} & \textbf{Layer 1} & \textbf{Layer 2} & \textbf{Layer 3} & \textbf{Layer 4} & \textbf{Layer 5} & \textbf{Layer 6} \\
\midrule
\textbf{a} &  -0.3847 & -0.5769 & -0.7609 &-1.0600 & -1.5256 & -1.6884 & -1.0240 \\
\textbf{b} &  3.3964 & 4.0830 & 4.7054 & 4.9473 & 4.7470 & 4.4954 & 4.1511 \\
\bottomrule
\end{tabular}%
}
\caption{Learned values for token weights in the HSM($a, b$) model.}
\label{tab:ab_layers}
\end{table}

\begin{table*}[t]
\centering
\includegraphics[width=\linewidth]{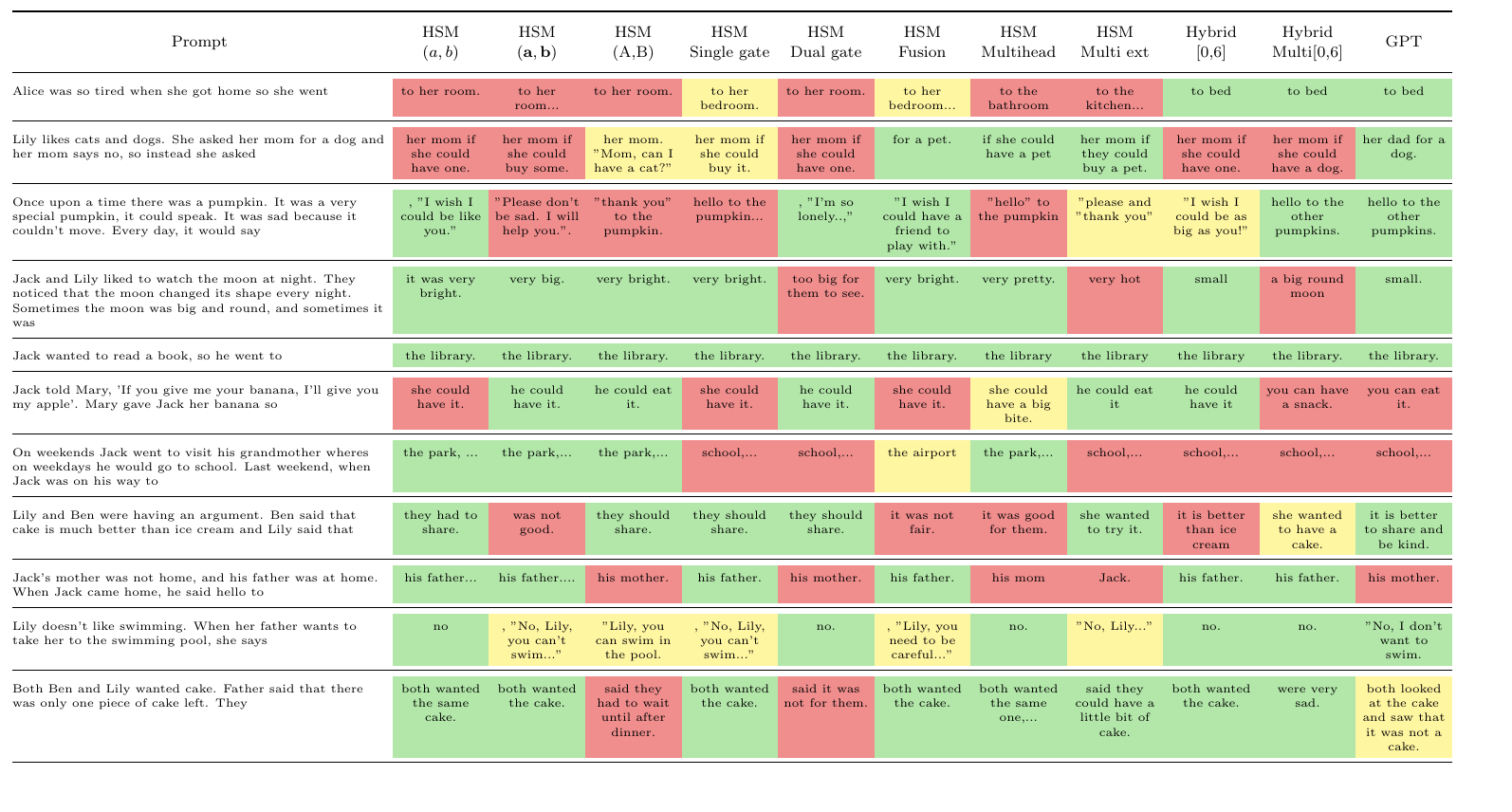} 
\caption{Qualitative performance using factual and reasoning prompts.}
\label{tab:subjective}
\end{table*}

\section{8. Discussions}
From Tables \ref{tab:objective} and \ref{tab:subjective} it can be seen that nearly all proposed HSM mixing functions perform more or less equally well. They produce grammatically correct sentences, and the differences relative to GPT are marginal, both in terms of validation loss (less than 10\% difference) and semantic quality. Of particular interest is that the simplest ($a, b$) function is the best-performing HSM variant. This indicates that the complex attention mechanism can be replaced by pairwise linear combinations, provided that the freed parameters are reallocated to the FFN component. \\

As noted earlier, replacing the first and last layers of GPT with those of HSM ($a, b$) Multihead, improves performance beyond the GPT baseline. One might expect that using the more powerful HSM ($a, b$) Multihead-ext version in such a hybrid configuration would yield further gains. However, this was not observed. One possible reason is that the additional shift permutations in this variant only affect the first and last layers, while the intermediate layers remain standard GPT layers.

\paragraph{Scaling}
To enable fair comparisons, all mixers were scaled to have the same number of learnable parameters. In principle, several degrees of freedom could be adjusted for this purpose, including the embedding dimensionality (\textit{dim}), context length, vocabulary size, number of layers and FFN dimensionality (\textit{hidden\_dim}). However, only one of these is suitable in the present setting, since our goal is to combine HSM with GPT layers within the stack. This requires identical data dimensionality and compatible parameter structures across layers. These constraints are satisfied only by scaling the FFN size as shown in Table \ref{tab:objective}.

\paragraph{Attention vs token mixing}
We have been careful not to use the term \textit{attention} unless we refer specifically to the attention mechanism described in \cite{vaswani2017attention}. This raises the question of what distinguishes our various token mixing functions from the attention concept? Does attention necessarily require the dense, dot-product-based, softmax-normalized weighting scheme introduced in the above Transformer? We argue that this is not strictly necessary, although certain properties must be present.
To clarify this distinction, we examine the proposed token mixing functions and assess which of them come close to implementing what could reasonably be described as "true attention".

We begin with the ($\mathbf{a}, \mathbf{b}$) scheme (noting that the ($a, b$) scheme is a special case). The operations performed are equivalent to depthwise one-dimensional convolutions with two taps. Stacking seven layers with shifts 1,2,4...64 results in a multi-timescale causal filter bank feeding the FFNs. There are no Q/K/V projections, dot products, or softmax normalization, and no content-dependent weighting. Context is introduced solely through fixed temporal shifts and learned per-channel scalars. While this scheme enables pairwise token interaction, it lacks several key properties of attention in the canonical Transformer sense, including content-based weighting, dense all-to-all mixing, and adaptive contextualization. The qualitative difference is that ($\mathbf{a}, \mathbf{b}$) uses fixed, content-independent mixing weights, whereas attention employs dynamic, content-dependent weights that adapt to the token values in each sequence.

The (A,B) scheme introduces learned linear projections, adding some resemblance to the query and key  mappings used in attention, but the mixing itself  remains content independent.
Moving on to the gated mixers, the single-input gated weighting introduces content dependence, although it depends only on one of the inputs ($\mathbf{x}$). It can be interpreted as a content-adaptive filter. 
The double-input gated weighting scheme fulfills the basic requirements for content-dependent interaction between token pairs and can reasonably be described as performing local pairwise attention. Finally, the fusion function is, in principle, capable of representing the dot-product computations and weighting operations required for attention. What all HSM variants ultimately lack, however, is the ability to compute normalized weights over all token pairs, as required for dense attention. This limitation is the fundamental trade-off that allows HSM to avoid quadratic-time complexity.

\section{9. Conclusion}
We have introduced \textbf{Hierarchical Shift Mixing (HSM)}, a general framework for efficient token mixing in Transformer-based language models. Rather than computing dense, content-normalized interactions between all token pairs within each layer, HSM distributes pairwise interactions hierarchically across layers, enabling linear-time complexity while preserving causality and parallelism.

We have shown that a wide range of HSM variants, spanning simple linear combinations, gated nonlinear mixers, and learned fusion mechanisms, achieve performance close to a GPT baseline. Notably, the simplest scalar-weighted ($a, b$) mixer consistently performs among the best HSM variants when model capacity is reallocated to the feed-forward networks. This suggests that the expressive power of Transformer models is not solely determined by the complexity of the attention mechanism, but also by how and where parameters are allocated within the architecture.

We further demonstrated that hybrid architectures combining HSM and GPT attention layers can outperform GPT while reducing training time. In particular, selectively replacing only a small number of attention layers with HSM layers yields a favorable trade-off between efficiency and predictive performance, even though such hybrids no longer maintain linear-time complexity.

As a final note, we stress that the models and training dataset used in this work are small compared to contemporary large language models. While the results are encouraging, further study is required to assess how the observed trends scale to larger models and datasets.

\section{10. Related Work}
Using hierarchical structures for signal processing is not a new idea. Before Transformers entered the scene, several hierarchical and multi-scale techniques were used to handle both mixing of samples and long sequences \cite{bahdanau2016neuralmachinetranslationjointly}, \cite{oord2016wavenetgenerativemodelraw}. Convolutional Neural Networks, arranged hierarchically, such as dilated CNNs, were strongly competitive at the time \cite{kalchbrenner2017neuralmachinetranslationlinear}, \cite{gehring2017convolutionalsequencesequencelearning}. From 2018  onward, the Transformer architecture with its attention mechanism became dominant, and dilated CNN language models became less central. Recently, however, convolution-based architectures have re-emerged, although in forms different from earlier approaches. This renewed interest primarily concerns convolution-like token mixing schemes that scale better than dense attention, often implemented as implicit or long convolutions combined with gating, rather than the classic WaveNet-style dilation schedule \cite{poli2023hyenahierarchylargerconvolutional}, \cite{peng-etal-2023-rwkv}, \cite{ding2023longnetscalingtransformers1000000000}, \cite{mai2023hypermixer}\\
\indent Current work on shift-based token mixing is  more focussed on vision models see e.g. \cite{liu2024hmhsa}, with \cite{peng2025rwkv7gooseexpressivedynamic} being a notable large language model exception. At the same time, the search for alternatives to quadratic-time dense attention has renewed interest in efficient token mixing schemes. The Hyena Hierarchy model \cite{poli2023hyenahierarchylargerconvolutional} computes long convolutions using the fast Fourier Transform (FFT) achieving $O(Tlog(T))$ complexity. Other recent work claims to approach dense attention performance with linear-time complexity \cite{yang2025gateddeltanetworksimproving}, \cite{kimiteam2025kimilinearexpressiveefficient}. \\
\indent The observation that hybrids of linear-time token mixing and dense attention can perform as well as, or even better than, pure dense attention was made shortly after the introduction of attention. After being relatively dormant for several years, this idea has recently entered state-of-the-art implementations. For example, it is incorporated in the KIMI Linear algorithm \cite{kimiteam2025kimilinearexpressiveefficient}. In addition, the GPU hardware manufacturer NVIDIA employs hybrid token mixing in its recent LLM architecture \cite{gu2025jetnemotronefficientlanguagemodel}.\\

\noindent \textbf{Acknowledgements}
The author is grateful to Prof. Onur Günlü for fruitful comments and corrections.

\bibliographystyle{ieeetr}
\bibliography{references}

\nocite{*} 

\newpage

\end{document}